\documentclass[letterpaper, 10 pt, conference]{ieeeconf}  
\IEEEoverridecommandlockouts                              
\usepackage{graphicx} 

\usepackage{amsmath} 
\usepackage{amssymb}  
\usepackage{booktabs}
\usepackage[dvipsnames]{xcolor}
\usepackage{paralist}
\usepackage{hyperref}
\usepackage{adjustbox}
\usepackage{cleveref}
\usepackage{multirow}
\usepackage{balance}
\usepackage{subfigure}

\crefname{equation}{\,}{\,}
\Crefname{equation}{Equation}{Equations}
\creflabelformat{equation}{(#1)}

\crefname{figure}{Fig.}{Figs.}
\Crefname{figure}{Figure}{Figures}

\crefname{table}{Tab.}{Tabs.}
\Crefname{table}{Table}{Tables}

\crefname{chapter}{Chap.}{Chaps.}
\Crefname{chapter}{Chapter}{Chapters}

\crefname{section}{Sec.}{Secs.}
\Crefname{section}{Section}{Sections}

\crefname{subsection}{Sec.}{Secs.}
\Crefname{subsection}{Section}{Sections}

\crefname{subsubsection}{Sec.}{Secs.}
\Crefname{subsubsection}{Section}{Sections}

\crefname{prob}{Prob.}{Probs.}
\Crefname{prob}{Problem}{Problems}

\crefname{property}{Prop.}{Props.}
\Crefname{property}{Property}{Properties}

\crefname{constr}{Constraint}{Constraints} 
\Crefname{constr}{Constraint}{Constraints}

\crefname{algocf}{Algorithm}{Algorithms}
\Crefname{algocf}{Algorithm}{Algorithms}

\crefname{algorithm}{Algorithm}{Algorithms}
\Crefname{algorithm}{Algorithm}{Algorithms}

\crefname{hyp}{Hyp.}{Hyps.}
\Crefname{hyp}{Hypothesis}{Hypothesis}

\crefname{ass}{Assumption}{Assumptions}
\Crefname{ass}{Assumption}{Assumptions}

\crefname{lem}{Lem.}{Lems.}
\Crefname{lem}{Lemma}{Lemma}

\crefname{prop}{Prop.}{Props.}
\Crefname{prop}{Proposition}{Propositions}

\newcommand{\red}[1]{{\color{black}{#1}}} 

\newcommand{\myparagraph}[1]{\vspace{2.0pt}\noindent\textbf{#1.}}

\title{\LARGE \bf
Learning Sequential Descriptors for\\ Sequence-based Visual Place Recognition
}

\author{Riccardo Mereu$^{*,1}$, Gabriele Trivigno$^{*,1}$, Gabriele Berton$^1$, Carlo Masone$^{2,1}$ and Barbara Caputo$^1$
\thanks{*These authors contributed equally.}
\thanks{$^{1}$ All authors are with the Visual and Multimodal Applied Learning Lab, Department of Control and Computer Engineering, Politecnico di Torino, 10138 Torino, Italy {\tt\small r.mereu@studenti.polito.it, \{gabriele.trivigno,gabriele.berton\}@polito.it}}%
\thanks{$^{2}$ Carlo Masone is also with Consorzio Interuniversitario Nazionale per l'Informatica (CINI),
        00185 Roma, Italy}%
}

\begin{document}

\maketitle
\thispagestyle{empty}
\pagestyle{empty}

\begin{abstract}
In robotics, Visual Place Recognition is a continuous process that receives as input a video stream to produce a hypothesis of the robot's current position within a map of known places. This task requires robust, scalable, and efficient techniques for real applications. 
This work proposes a detailed taxonomy of techniques using sequential descriptors, highlighting different mechanism to fuse the information from the individual images. This categorization is supported by a complete benchmark of experimental results that provides evidence on the strengths and weaknesses of these different architectural choices. In comparison to existing sequential descriptors methods, we further investigate the viability of  Transformers instead of CNN backbones, and we propose a new ad-hoc sequence-level aggregator called SeqVLAD, which outperforms prior state of the art on different datasets. The code is available at \url{https://github.com/vandal-vpr/vg-transformers}
\end{abstract}

\section{INTRODUCTION}

\begin{figure}[t!]
  \centering%
  \includegraphics[width=0.9\columnwidth]{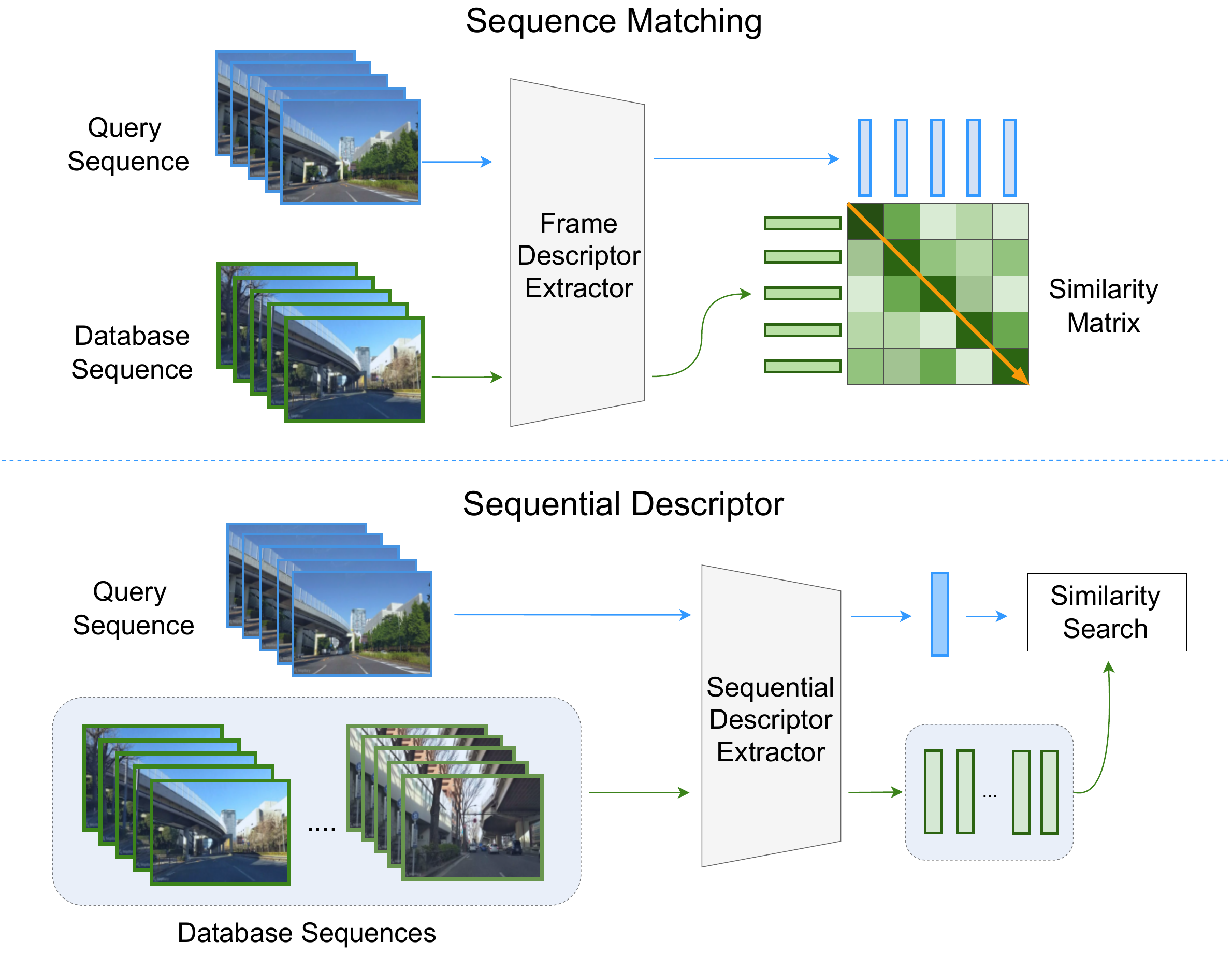}
  \caption{(Top) Sequence matching individually processes each frame in the sequences to extract single-image descriptors. The frame-to-frame similarity scores build a matrix, and the best matching sequence is determined by aggregating the scores in the matrix. (Bottom) With sequential descriptors, each sequence is mapped to a learned descriptor, and the best matching sequence is directly determined by measuring the sequence-to-sequence similarity.}
  \label{fig:teaser}
  \vspace{-0.4cm}
\end{figure}

In mobile robotics, recognizing previously seen places, i.e., Visual Place Recognition (VPR), is important both to perform loop closure in SLAM  and as a localization alternative to GPS.
This task is configured as a continuous process that consumes a stream of images captured by the robot as it navigates the environment, which raises the question of how to leverage the temporal information in the data most effectively.
The predominant approach to exploit this sequential information, popularized by SeqSLAM~\cite{seqslam}, is to perform \emph{sequence matching} (see \cref{fig:teaser} top). First, each frame of the input sequence is individually compared to the collection of images of known places (referred to as database) to build a similarity matrix. Then, this matrix is searched for the most likely trajectory by aggregating the similarity scores.
Albeit effective, this methodology has some drawbacks. On the one hand, the mechanism of searching through the similarity matrix relies heavily on assumptions of homogeneity regarding the motion model of the robot, making it difficult to generalize to other conditions. On the other, as remarked in~\cite{Garg-2021-seqNet}, this solution is inefficient because its computational cost grows linearly with both the sequence length and the size of the map.

Few recent works have recognized these limitations of sequence matching and have independently proposed to use \emph{sequential descriptors} that summarize sequences as a whole, thus enabling to directly perform a sequence-to-sequence similarity search~\cite{Garg-2021-seqNet, Facil-2019-multiViewNet, Warburg-2020-msls} (see \cref{fig:teaser} bottom).
This idea is promising because i) it is more efficient and scalable than sequence matching, and ii) a sequential descriptor naturally captures the temporal information in the video stream, which provides more robustness to high-confidence false matches than single image descriptors.
However, research on sequential descriptors is still in its infancy, and so far there is no detailed comparative study on the impact of the different architectural choices.
To fill this void, we propose a detailed investigation that makes the following contributions:
\begin{itemize}
  \item an exhaustive taxonomy for sequential descriptors methods, with the aim of better understanding how different methods relate to each other;
  \item a thorough benchmark of experimental results, to understand how different methods rank in terms of recall@N, but also to investigate their scalability, hardware requirements, and overall practicality in real-world scenarios on small and large-scale datasets;
  \item we investigate the use of Transformers for sequence-based VPR, whereas all previous methods rely on CNN backbones;
  \item we design a new aggregation layer, called \emph{SeqVLAD}, that exploits the temporal cues in a sequence and leads to a new state of the art on multiple datasets, outperforming the second-best results on all datasets.
\end{itemize}


\section{RELATED WORKS}
Visual Place Recognition from sequences of frames~\cite{Lowry-2016, Masone-2021} is tackled mainly by two families of algorithms: sequence matching and sequential descriptors.

\myparagraph{Sequence matching}
Sequence matching is a well established paradigm \cite{first-seq-search,seqslam} that operates in two steps. First, a similarity matrix is built by sequentially comparing the single image descriptors of each frame in the query with those of all the frames in the database sequences. Then, the best fitting database sequence is determined by aggregating the individual similarity scores under simplifying assumptions, such as a constant velocity of the robot, absence of loops within the sequences, or no stops \cite{vpr-hard}, which makes it hard to generalize to real-world applications. Many works have focused on relaxing these assumptions and robustifying the method, e.g., by using difference-based representations \cite{Garg-2020-deltaDescr}, exploiting odometry information \cite{slam-odometry, Naseer2014-odometry}, querying frames by their relative distance to allow for traversals at different speeds \cite{Pepperell-2014}, using more complex search methods~\cite{Hansen-2014, Naseer-2018}.
Recently, SeqMatchNet \cite{Garg2021SeqMatchNetCL} has addressed  the fact that these methods rely on learned single image descriptors trained without considering the downstream procedure of score aggregation.
Nevertheless, there are other intrinsic drawbacks in sequence matching~\cite{Garg-2021-seqNet}: i) misleading single image descriptor matches may cause false positives, which could be instead detected from the sequential information in the sequence, and ii) the computational cost of the matching scales linearly with database size and sequence length.

\myparagraph{Sequential descriptors}
Sequential descriptor methods summarize each sequence with a single descriptor and then perform the similarity search directly sequence-by-sequence. 
This allows to reduce the cost of the matching and to incorporate temporal clues into the descriptors.
Although this idea has been explored in fields marginally related to VPR, such as Video Re-Identification \cite{sequential-vlad, timesformer, personvlad} and localization from 3D measurements \cite{pointnetvlad,3d-vl}, only a handful of works have explicitly adopted it for VPR.
Facil et al. \cite{Facil-2019-multiViewNet} first introduced the idea of sequential descriptors in VPR using three basic techniques: concatenation of single image descriptors, fusion of the frame-level features with an FC layer, and integration over time of the single-image features via an LSTM network. Some of these results are further extended in \cite{Warburg-2020-msls} on the Mapillary Street Level Sequences (MSLS) dataset.
More recently, SeqNet \cite{Garg-2021-seqNet} proposed to use a 1D temporal convolution to perform a learned pooling of frame-level features into a sequential descriptor. However, this method is implemented on top of pre-computed single image descriptors, which impedes fine-tuning the single frames' representation. Additionally, the implementation of SeqNet descriptors is not readily scalable to train on large datasets because it requires loading in memory the entire set of single-image descriptors.

In comparison to these prior works \cite{Garg-2021-seqNet,Facil-2019-multiViewNet,Warburg-2020-msls}, in this paper we consider a more broad investigation of different architectures to learn sequential descriptor, resulting in a comprehensive taxonomy of these methods. Additionally, within this investigation we propose novel architectural solutions, such as the use of Transformer-based architectures and the first aggregation layer specific for sequence-based VPR.


\section{METHODOLOGY}
\label{sec:methodology}
In this section, we first formalize the problem of sequential VPR, then we introduce a taxonomy of architectures for learning to extract sequential descriptors.

\subsection{Problem Setting}
We define the sequential VPR task following the \emph{seq2seq} setting introduced by~\cite{Warburg-2020-msls}.
Formally, given a query sequence to be localized, the task is posed as a retrieval problem where the result is deemed correct if at least one sequence within the top-N results correctly matches the query (recall@N metric). By correct match, it is intended that any of the frames in the query sequence is less than 25 meters away from any of the frames in the database sequence.

Our goal is to investigate how to build sequential descriptors to tackle this task. Namely, given a sequence $I^L$ of $L$ images, we seek a learnable mapping $f_\theta: \; I^L \rightarrow \mathbb{R}^D$ that produces a single $D$-dimensional descriptor that represents the sequence as a whole. 
Throughout our work we focus on place recognition with \textit{short sequences}, as previously done by \cite{Garg-2021-seqNet, Garg2021SeqMatchNetCL}, targeting \textit{accurate, efficient} place recognition, thus avoiding delays that processing long sequences would cause, not acceptable in practical applications.  
However, we also provide an ablation study on the effect of varying the sequence length $L$ at test time (cf. \cref{sec:experiments}).

\begin{figure*}[t!]
  \centering
  \includegraphics[width=\textwidth]{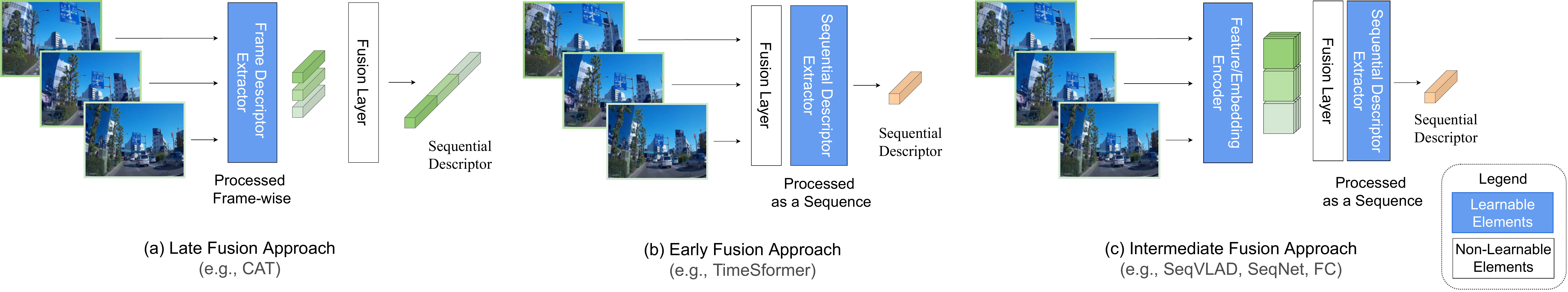}
  \caption{\red{Taxonomy of the sequential descriptors architectures, obtained by composing learnable networks and parameterless fusion.}
  }
  \label{fig:architectures}
  \vspace{-0,4cm}
\end{figure*}

\subsection{Architectures for Learned Sequential Descriptors}
\label{sec:sequential_descriptors}
All previous works \cite{Garg-2021-seqNet,Facil-2019-multiViewNet,Warburg-2020-msls} combine learnable functions with some form of parameterless fusion, e.g., concatenation of vectors, to learn a mapping $f_\theta: \; I^L \rightarrow \mathbb{R}^D$. 
Here we propose to categorize these architectures depending on the stage where the fusion mechanism is applied. Following this criterion, we identify three paradigms, which are depicted in \cref{fig:architectures}: late fusion, early fusion, and intermediate fusion. Hereinafter, we describe these categories and discuss not only how existing methods fit them, but we also introduce new methods within each category.

\subsubsection{\textbf{Late fusion}}
In late fusion methods there is a trainable network that processes individually each image in the sequence, and only at the end these separate flows are fused to create a single sequential descriptor (see \cref{fig:architectures}a).
This paradigm is the easiest to implement, since it can leverage existing models for processing single images. However, this simplicity comes at a cost, because even if the whole architecture is fine-tuned using a loss defined on the final sequential descriptor, this structure imposes that the trainable network can only see one frame at a time, thus limiting its access to the temporal information.
Most of the existing methods in literature fall in this category \cite{Facil-2019-multiViewNet,Warburg-2020-msls}, using already available convolutional networks to process single images, such as NetVLAD \cite{Arandjelovic-2018} and GeM \cite{Radenovic-2019}, and fusing  their outputs via concatenation or pooling.

We expand upon these existing methods by considering more modern architectures based on visual Transformers, which process the image as a sequence of tokens. Specifically, we experiment with two different visual Transformers as backbones: ViT~\cite{vit} and CCT~\cite{cct}.
ViT takes a
an additional learnable token, referred to as \emph{CLS token} whose corresponding output embedding has been proven to be an effective image representation for downstream tasks as classification \cite{vit} or image retrieval \cite{ElNouby-2021}. We use this embedding as a single image vector representation.
 On the other hand, CCT~\cite{cct} relies on an ad-hoc pooling layer named SeqPool, which takes the output embeddings corresponding to all the input tokens and produces a single vectorial representation of the image. 
In both cases, we perform the fusion of the single image embeddings by using concatenation.

\subsubsection{\textbf{Early fusion}}
At the other end of the spectrum, the idea of early fusion is to have a learnable model that directly processes all the frames at once (see \cref{fig:architectures}b). This gives the model the flexibility to access all the temporal information in the sequence, but such an unconstrained network may require more resources to be trained. 
To the best of our knowledge, at the moment of this writing there are no existing sequential descriptor methods that implement this strategy. As a first solution of this kind, we propose to use the TimeSformer~\cite{timesformer}, a model originally developed for the task of Action Recognition and based on a self-attention scheme that extends to both the spatial and temporal axes. 
The TimeSformer treats the input frames as a set of non-overlapping patches, which are then flattened and mapped to a set of tokens via a learnable matrix.
Therefore, in this case the sequential patchification and flattening of the input sequence is a fusion operation that directly presents the network with information from all the frames.
As done in ViT \cite{vit}, the TimeSformer adds a CLS token to the set of input tokens. We use its output embedding as the sequential descriptor.

\subsubsection{\textbf{Intermediate fusion}}
Intermediate fusion is a tradeoff between early and late fusion. In this paradigm there is a first learnable stage that processes individual frames, then a fusion operation that collects all the flows, and finally another learnable element that outputs the sequential descriptor (see \cref{fig:architectures}c).
This strategy provides more flexibility w.r.t. late fusion, because the trainable model after the fusion can access information from all the frames. 
At the same time, the access to the temporal information from all frames is more constrained than in the case of the early fusion and can lead to less resource-demanding models.
Both \cite{Facil-2019-multiViewNet} and \cite{Garg-2021-seqNet} demonstrate examples of this paradigm, with a first convolutional network that extracts frame-level features, a flattening and reshaping operation that fuses them, and a final learnable layer that generates the sequential descriptor (a FC layer in \cite{Facil-2019-multiViewNet} and a 1D temporal convolution in SeqNet~\cite{Garg-2021-seqNet}).

\begin{figure}[t!]
  \centering
  \includegraphics[width=\columnwidth]{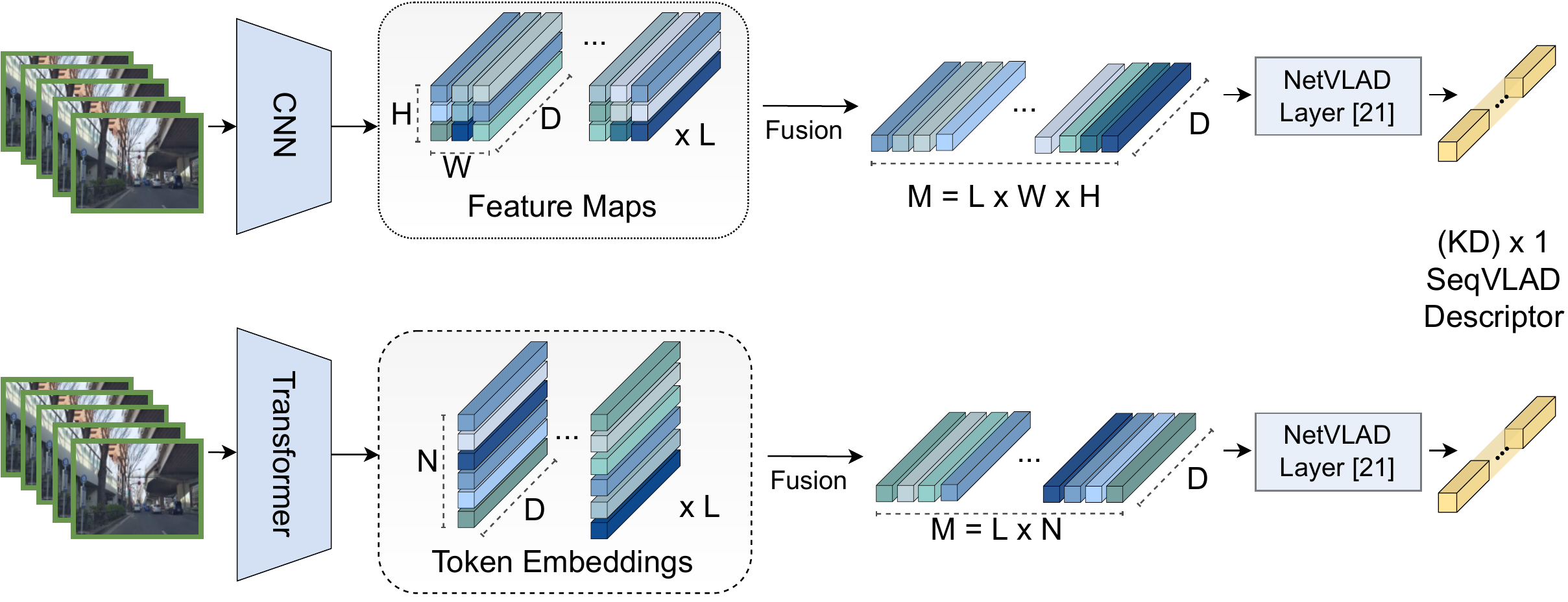}
  \caption{Representation of the SeqVLAD layer, in conjunction with a CNN backbone (top) and a Transformer backbone (bottom).}
  \label{fig:seqvlad}
  \vspace{-0,3cm}
\end{figure}

Here we propose a new intermediate fusion method, named \emph{SeqVLAD}, that is designed to generalize the idea of NetVLAD~\cite{Arandjelovic-2018} in order to handle sequences of frames of arbitrary length.
We recall that the intuition of NetVLAD is to i) interpret the $H \times W \times D$ tensor of feature maps of a CNN as a set of $(H \cdot W)$ $D$-dimensional local, densely extracted descriptors, ii) cluster them in terms of $K$ \textit{visual words} \cite{Jegou-2011, Arandjelovic-2018}, and iii) use as global descriptor a statistic of the obtained cluster-assignment.
In order to handle multiple frames, SeqVLAD prepends the NetVLAD layer with a reshaping operation (fusion) that interprets all the frame-level features as a collection of $M$ $D$-dimensional local descriptors. 
We propose two different implementations of this fusion operation to handle both the case of convolutional and Transformer-based backbones (see \cref{fig:seqvlad}).

\paragraph{SeqVLAD for CNNs} given the input sequence $I^L$, the CNN backbone  produces a tensor of shape $L \times H \times W \times D$, with $L$ equal to the sequence length, and $H \times W \times D$ being the shape of the features maps of each frame. SeqVLAD interprets this tensor as a set of $M$ $D$-dimensional local feature descriptors with $M = L \cdot H \cdot W$.

\paragraph{SeqVLAD for Transformers} given the input sequence $I^L$, a Transformer backbone outputs a set of $T$ $D$-dimensional embeddings for each frame. SeqVLAD sees this output as a set of $M$ $D$-dimensional embeddings, with $M = L \cdot T$.

After this fusion operation, as in the classical NetVLAD~\cite{Arandjelovic-2018}, the $M$ embeddings are aggregated using a soft-assignment of their residuals w.r.t. the centroid of their corresponding cluster, i.e.,
\begin{equation}
    \label{eq:SeqVLAD}
    \textit{SeqVLAD}(k) = \sum_{f=1}^{L} \sum_{i=1}^{M} \bar{a}_k(\mathbf{x}_{fi}) \cdot (\mathbf{x}_{fi} - \mathbf{c}_k)
\end{equation}
where $x_{fi}$ is a single feature or embedding, $\mathbf{c}_k$ is the $k$-th centroid, and  $\bar{a}_k(\mathbf{x}_{fi})$ is the soft-assigment defined as
\begin{equation}
    \label{eq:soft_assignment}
    \bar{a}_k(x_{fi}) = \frac{e^{\mathbf{w}^T_k x_{fi} + b_k}}{ \sum_{k'=1}^{K}  e^{\mathbf{w}^T_{k'} x_{fi} + b_{k'} } }
\end{equation}

Like the other existing intermediate fusion methods \cite{Garg-2021-seqNet,Facil-2019-multiViewNet}, SeqVLAD's learned mapping is aware of the temporal information in the sequence. However, differently from \cite{Garg-2021-seqNet,Facil-2019-multiViewNet} which use general purpose learnable layers to capture this sequential information, SeqVLAD extends an aggregation layer that was specifically designed for VPR. Additionally, SeqVLAD has other noteworthy characteristics:
\begin{itemize}
  \item \textit{Scalability}: the number of parameters and the output dimensionality does not depend on the sequence length;
  \item \textit{Flexibility}:  being agnostic to the length of the sequence,  it allows sequences of different lengths to be comparable with each other. This characteristic makes SeqVLAD an extremely practical method for real-world applications, given that the sequences used in various stages of product development might have different lengths.
\end{itemize}

\section{Experiments} \label{sec:experiments}
This section presents the experimental setup used and the results obtained throughout our investigation of sequence-based visual place recognition methods.

\subsection{Experimental setup}
\label{sec:exp_setup}

\myparagraph{Datasets} 
For our experiments we use two datasets: MSLS~\cite{Warburg-2020-msls} and Oxford RobotCar \cite{Maddern20171Y1}, whose statistics are summarized in \cref{tab:datasets}. 

MSLS is a recent dataset that was built purposely  to support research on large scale VPR and that comprises a heterogeneous collection of sequential street-level images gathered from various cities around the world and in multiple environmental conditions. 
Unfortunately, the authors of MSLS never released the ground truths for the test split, nor have announced plans to do it.
To circumvent this issue many works, following \cite{Hausler-2021}, report recalls computed directly on the validation set.
We argue that for a fairer evaluation it is necessary to adopt a proper train-validation-test split and we propose the following one:
\begin{itemize}
    \item \textit{MSLS Test set}: \textit{Copenaghen, San Francisco}: the former official validation set;
    \item \textit{MSLS Validation set}: \textit{Amsterdam, Manila}: these two cities were released in~\cite{Warburg-2020-msls} as part of the training set. We used as criteria to choose them (i) size comparable to the test split (ii) use two cities, like the formerly proposed validation set, from two different continents;
    \item \textit{MSLS Train set}: same as the official one from~\cite{Warburg-2020-msls}, with the exception of the 2 cities chosen for validation.
\end{itemize}
Given the scale of the MSLS training set (cf. \cref{tab:datasets}), not all methods are practically trainable on it (this point is discussed in \cref{sec:results}). In order to compare the results with less scalable methods we also conduct another set of experiments where the models are trained on a single city from MSLS dataset. To comply with the experimental setup of \cite{Garg-2021-seqNet, Garg2021SeqMatchNetCL}, we use Melbourne for this purpose, while retaining the same validation and test sets defined above.

Oxford-Robotcar \cite{Maddern20171Y1} contains sequences collected through multiple traversals along a single route in the city of Oxford, but repeated in different seasons and environmental conditions.
Despite its smaller scale and size w.r.t. MSLS (cf. \cref{tab:datasets}), Oxford-Robotcar allows us to 
investigate the robustness of the tested methods to the Day-Night domain shift and how they are affected by a smaller training dataset.

Since a consistent split of Oxford-RobotCar sequences is lacking in previous literature \cite{Garg-2020-deltaDescr, Garg-2021-seqNet, Garg2021SeqMatchNetCL}, we adopt the following lap split:
\begin{itemize}
	\item \textit{RobotCar Test set}:
	queries:  2014-12-16-18-44-24 (winter night); 
    database:  2014-11-18-13-20-12 (fall day).
	\item \textit{RobotCar Validation set}:
	queries:  2015-02-03-08-45-10 (winter day, snow);
    database:  2015-11-13-10-28-08 (fall day, overcast).
    \item \textit{RobotCar Train set}: 
    queries:    2014-12-17-18-18-43 (winter night, rain);
	database:  2014-12-16-09-14-09 (winter day, sun).
\end{itemize}

\myparagraph{Methods} 
We perform experiments with methods from all three paradigms described in \cref{sec:methodology}. 
Regarding late fusion approaches, we test the concatenation (CAT) of NetVLAD and GeM frame-descriptors~\cite{Warburg-2020-msls} and we also consider the concatenation of frame-level representations from more recent Transformer-based architectures, namely ViT~\cite{vit} and CCT~\cite{cct} (in two variants, CCT224 and CCT384).
For intermediate fusion, we include the results obtained with SeqNet \cite{Garg-2021-seqNet} as well as experiments using a Fully-Connected (FC) layer on the flattened and combined (fused) frame-level representations extracted with NetVLAD and with ViT.
We also show the results achieved with SeqVLAD using a variety of different backbones, both CNNs and Transformers.
Finally, as an early fusion technique we use the TimeSformer \cite{timesformer} architecture.
It is important to point out that the Transformer-based models need to work with resized frames:  TimeSformer, ViT and CCT224 work with images resized to $224 \times 224$, whereas CCT384 accepts a resolution of $384 \times 384$. For CNNs we used a resolution of $480 \times 640$. 
For completeness, we also report the results achieved with sequence matching methods, including a popular approach like SeqSLAM \cite{seqslam} and more recent state-of-the-art architectures such as HVPR \cite{Garg-2021-seqNet}, SeqNet \cite{Garg-2021-seqNet}, Delta Descriptors~\cite{Garg-2020-deltaDescr} and SeqMatchNet~\cite{Garg2021SeqMatchNetCL}.
In order to make results comparable, we apply the PCA dimensionality reduction in order to obtain a constant descriptor dimensionality. In practical applications it is also useful as a mean of speeding up the retrieval thanks to smaller descriptors.
The PCA is computed only after the training is complete, and it is therefore only used for inference.
For the baseline methods SeqSLAM \cite{seqslam}, Delta Descriptors \cite{Garg-2020-deltaDescr}, SeqNet and HVPR \cite{Garg-2021-seqNet}, and SeqMatchNet \cite{Garg2021SeqMatchNetCL} we used their official implementations.

\myparagraph{Training}
\label{sec:training}
To train the models we follow the protocol defined in \cite{Warburg-2020-msls}, which adapts the triplet loss training from NetVLAD \cite{Arandjelovic-2018} to sequential scenarios.
This contrastive setting heavily depends on a careful selection of hard negatives. We use the mining methods from \cite{Warburg-2020-msls}, which relies on a cache with 1000 randomly sampled negative sequences to avoid the computation of the features of the whole database. 
This cache is refreshed after iterating over 1000 triplets.
Following \cite{Warburg-2020-msls}, we applied early stopping by interrupting the training if the recall@5 on the validation set does not improve for 5 epochs.
We define a training epoch as a pass over 5,000 queries, and we use the Adam optimizer \cite{Kingma-2014} for training.
We use a batch size of 4 triplets,
where each triplet is composed of a query sequence, its positive, and 5 negatives.
Finally, at inference time the candidates are retrieved through an exhaustive kNN search, as in \cite{Warburg-2020-msls}.

\begin{table}[t!]
  \centering
    \caption{
    \red{Number of sequences composing the datasets.}
    }
    \vspace{-0,1cm}
  \label{tab:datasets}
  \resizebox{\columnwidth}{!}{
    \begin{tabular}{@{}cccccc@{}}
    \toprule
    Dataset Name &
      \begin{tabular}[c]{@{}c@{}}Image\\Resolution\end{tabular} &
      \begin{tabular}[c]{@{}c@{}}Seq.\\Length\end{tabular} &
      \begin{tabular}[c]{@{}c@{}}\# train\\ db/queries\end{tabular} &
      \begin{tabular}[c]{@{}c@{}}\# validation\\ db/queries\end{tabular} &
      \begin{tabular}[c]{@{}c@{}}\# test\\ db/queries\end{tabular} \\ \midrule
    \multirow{3}{*}{MSLS} & \multirow{3}{*}{480$\times$640} & 5  & 733k / 393k & 8.1k / 5.8k & 13.6k / 8k \\
                          && 10 & 555k / 303k & 5.2k / 4.1k & 8.2k / 4.5k \\ 
                          && 15 & 478k / 259k & 4k   / 3.3k & 6.1k / 3.3k \\
    \midrule
    Melbourne         & 480$\times$640   & 5  & 95k / 76k & 8.1k / 5.8k & 13.6k / 8k \\
    \midrule
    Oxford RobotCar & 1280$\times$960    & 5  & 3.6k / 3.3k & 4k / 3.7k & 3.6k / 3.9k  \\ \bottomrule
    \end{tabular}}
\vspace{-0.5cm}
\end{table}

\subsection{Results and Discussion}
\label{sec:results}

\begin{table*}[ht!]
\caption{
Evaluation of sequential descriptors and sequence matching. 
SL stands for Sequence Length, CAT indicates concatenation of descriptors, FC stands for Fully Connected layer. {\color{blue}**} denotes a non-trained method. * refers to a single descriptor.}
\vspace{-0,4cm}
\label{tab_all}
\begin{center}
\resizebox{\textwidth}{!}{
\begin{tabular}{lccccccc}
\toprule
\multirow{2}{*}{\begin{tabular}[c]{@{}c@{}}Category\end{tabular}}&
\multirow{2}{*}{\begin{tabular}[c]{@{}c@{}}Method\end{tabular}} &
\multirow{2}{*}{\begin{tabular}[c]{@{}c@{}}Backbone\end{tabular}} &
\multirow{2}{*}{\begin{tabular}[c]{@{}c@{}}Descriptor Dimension\end{tabular}} &
\multirow{2}{*}{\begin{tabular}[c]{@{}c@{}}GPU Memory\\Occupation (GB)\end{tabular}}&
\multirow{2}{*}{\begin{tabular}[c]{@{}c@{}}Train on\\ Melbourne (R@1)\end{tabular}}&
\multirow{2}{*}{\begin{tabular}[c]{@{}c@{}}Train on\\ MSLS (R@1)\end{tabular}} &
\multirow{2}{*}{\begin{tabular}[c]{@{}c@{}}Train/Test on\\ Oxf.RobotCar (R@1)\end{tabular}}\\
&&&&&&&\\
\midrule
\multirow{3}{*}{\begin{tabular}[c]{@{}c@{}}Sequence\\Matching\end{tabular}} 
&SeqSLAM{\color{blue}**} \cite{seqslam}                         & VGG-16    & 4096$^*$& 2.68  & 45.9 &  \underline{45.9} & 34.7 \\
&HVPR \cite{Garg-2021-seqNet}                    & VGG-16    & 4096$^*$ & 2.68  & \underline{51.0} & - & \underline{56.8} \\
&SeqMatchNet\cite{Garg2021SeqMatchNetCL}         &VGG-16     & 4096$^*$ & 2.68  & 44.8 & - & 51.9 \\
\midrule\midrule
\multirow{10}{*}{\begin{tabular}[c]{@{}c@{}}Late\\Fusion\end{tabular}}
&Delta Descriptors{\color{blue}**}\cite{Garg-2020-deltaDescr}  & VGG-16    & 4096$^*$ & 2.68  &43.0& 43.0 & 18.0\\
& GeM + CAT \cite{Warburg-2020-msls}              & ResNet-18 & 1280     & 2.04  & 66.7 & 76.8 & 75.4\\
& GeM + CAT \cite{Warburg-2020-msls}              & ResNet-50 & 5120     & 2.25  & 63.4 & 68.6 &81.3 \\
& NetVLAD + CAT\cite{Warburg-2020-msls}           & ResNet-18 &16384  $\cdot$ SL   & 2.04  & 74.3 & 84.3 & 80.3\\
& NetVLAD + CAT\cite{Warburg-2020-msls}           &ResNet-50 &65536 $\cdot$ SL    & 2.26  & 73.3 & \underline{85.6} &\underline{92.1}\\
& \textbf{NetVLAD + CAT + PCA}                    & ResNet-18 & 4096     & 2.04  & \underline{75.5} & 83.7 & 67.8  \\
& \textbf{NetVLAD + CAT + PCA}                    &ResNet-50 &4096       & 2.26  & 74.9 & 85.3 & 89.3 \\
& \textbf{CLS + CAT}                              & ViT   &768 $\cdot$ SL& 2.19  & 68.2 & 78.1 & 69.8\\
& \textbf{SeqPool + CAT}                          & CCT224&384 $\cdot$ SL& 1.91  & 65.7 & 74.2 & 67.6\\
& \textbf{SeqPool + CAT}                          & CCT384&384 $\cdot$ SL& 2.01  & 69.9 & 77.8 & 77.4\\
\midrule
\multirow{15}{*}{\begin{tabular}[c]{@{}c@{}}Intermediate\\Fusion\end{tabular}}
&SeqNet \cite{Garg-2021-seqNet}                  & VGG-16    & 4096     & 2.68  & 50.1 & - & 60.5\\
&\red{\textbf{SeqNet}} \cite{Garg-2021-seqNet}                  & ResNet-18    & 4096     & 2.04  & 45.2 & - & 31.3\\
&\red{\textbf{SeqNet}} \cite{Garg-2021-seqNet}                  & ResNet-50    & 4096     & 2.26  & 45.6 & - & 61.3\\
&\red{\textbf{SeqNet}} \cite{Garg-2021-seqNet}                  & CCT224    & 4096     & 1.91  & 45.6 & - & 41.3\\
&\red{\textbf{SeqNet}} \cite{Garg-2021-seqNet}                  & CCT384    & 4096     & 2.02  & 45.4 & - & 58.8\\
&\textbf{CLS  + FC}                              & ViT       & 4096     & 2.25  & 66.8 & 78.1 &67.8 \\
&\textbf{NetVLAD + FC}                           & ResNet-18 & 4096     & 3.39  & 55.5 & 68.5 & 44.7 \\
&\textbf{NetVLAD + FC}                           & ResNet-50 & 4096     & 7.63   &  - & -& - \\ 
&\textbf{SeqVLAD + PCA}                          & ResNet-18 & 4096     & 2.04  & 78.2 & 85.5 & 86.5\\
&\textbf{SeqVLAD + PCA}                          & ResNet-50 & 4096     & 2.26  & 74.5 & 85.2 & 85.9 \\
&\textbf{SeqVLAD + PCA}                          & VGG-16    & 4096     & 2.68  & 78.5 & 85.8 & 79.8 \\
&\textbf{SeqVLAD + PCA}                          & ViT       & 4096     & 2.19  & 71.9 & 84.0 & 67.7 \\ 
&\textbf{SeqVLAD + PCA}                          & CCT224    & 4096     & 1.91  & 75.8 & 85.5 & 69.6 \\
&\textbf{SeqVLAD}                                & CCT384    &24576     & 2.02  & \textbf{\underline{81.7}} & \textbf{\underline{89.4}} & 92.8 \\
&\textbf{SeqVLAD + PCA}                          & CCT384    & 4096     & 2.02  & 81.4 & 89.2 &\textbf{\underline{93.3}} \\
\midrule
\multirow{2}{*}{\begin{tabular}[c]{@{}c@{}}Early Fusion\end{tabular}} &\red{\textbf{ResNet-50 2D +1}}             & - & 1024 & 3.05 & 63.9 & 75.2& \underline{80.9}\\
&\textbf{TimeSformer}                            & -         &  768     & 2.34  & \underline{73.8} & \underline{81.5} & 74.9 \\
\bottomrule
\end{tabular}}
\end{center}
\vspace{-0.5cm}
\end{table*}

\myparagraph{Sequential descriptors vs. sequence matching}
Tab. \ref{tab_all} presents an exhaustive comparison of all the methods, both on MSLS and Oxford-Robotcar.
The results show that sequence matching approaches systematically achieve lower results across the different settings compared to sequential descriptors. On the Melbourne split the best performing sequence matching technique is HVPR \cite{Garg-2021-seqNet}, which has a first filtering stage based on sequential descriptors and achieves a 51.0\% R@1 on the test set. The best sequential descriptor method, CCT384+SeqVLAD, obtains 81.4\%, with a considerable increase of 30\% in performance.
We also find that training sequence matching methods do not scale to large datasets such as MSLS due to computational constraints. For instance, SeqNet would require more than 1 TB of memory to train on the whole MSLS train set. We report with a dash "-" in \cref{tab_all} the results that could not be computed because of these limitations. This observation does not apply to  SeqSLAM and Delta Descriptors (marked with {\color{blue}**}) because they do not have a training phase and are only tested on the datasets.

\myparagraph{Experimental comparisons of sequential descriptors methods}
Tab. \ref{tab_all} shows that the ability to exploit the whole training set of MSLS, instead of the subset from Melbourne, is rather beneficial for obtaining models that generalize better on the test set (9\% increase on average), proving that train-time scalability is a factor that should be considered when training a sequence VPR model or proposing a novel training procedure.
Regarding the different groups of sequential descriptors architectures, we draw the following conclusions:
\begin{enumerate}
    \item
    The simple concatenation of single-frame descriptors (i.e., CAT) achieves better performance than their linear projection with a learnable Fully Connected layer (FC). The performance gap could be due to the significant number of learnable parameters that are inherently required when handling descriptors with a high dimensionality as input, as in the case of NetVLAD. This issue makes this approach more vulnerable to overfitting, which is especially evident when training on the Robotcar dataset due to its small size. FC layers also become impractical when the size of the descriptors grows due to the large computational resources required, e.g., a ResNet-50 + NetVLAD + FC would require 240 GB of GPU memory for training.
    \item SeqVLAD achieves the best results w.r.t. other methods, regardless of the dataset and the backbone. This confirms the intuition that a sequence-aware lightweight learnable layer is key for success in sequence-based VPR.
    We also note that applying PCA to SeqVLAD descriptors dramatically reduces the dimensionality with a negligible drop in recalls, allowing SeqVLAD to outperform the previous state-of-the-art with compact descriptors. Note that descriptors' dimensionality is a critical factor for sequence-based VPR, as thoroughly explained in the following paragraphs.

    \item The TimeSformer provides the only early-fusion architecture and achieves acceptable results with very compact descriptors, an essential characteristic for real-time applications (see next paragraph).
    Perhaps the biggest drawback of purely Transformers-based architectures (i.e., TimeSformer and ViT) is the quadratic relationship between the resolution of the input images and their computational complexity. This limits their use to small frames ($224 \times 224$ in our experiments), which could explain the non-optimal results achieved by these architectures.
    However, we believe that the comparable results with such low resolution and compact descriptor (768) w.r.t CNNs prove the validity of the early fusion concept, and making it more efficient could be a promising direction for future research.
    \item The choice of the backbone has a big influence on results, and we find that efficiency-oriented architectures such as ResNets \cite{He_2016_resnet} and CCTs achieve on average the best results,
    making this manuscript the first work to investigate this extremely important factor in sequence-based VPR, and proving that the use of the much more costly VGG-16 \cite{Simonyan_2015_vgg}, common in the literature, is not justified by its results.
    \item Methods relying on low ($224 \times 224$) resolution suffer from heavy recall drops when performing cross-domain VPR. This is noticeable from the results on RobotCar, where the test set presents a day/night shift. This could also be due to the small size of the training data, as generally Transformers are more data-hungry than CNNs since they lack their inductive bias 
\end{enumerate}

\begin{figure}[t!]
  \vspace{-0,1cm}
  \includegraphics[width=\columnwidth]{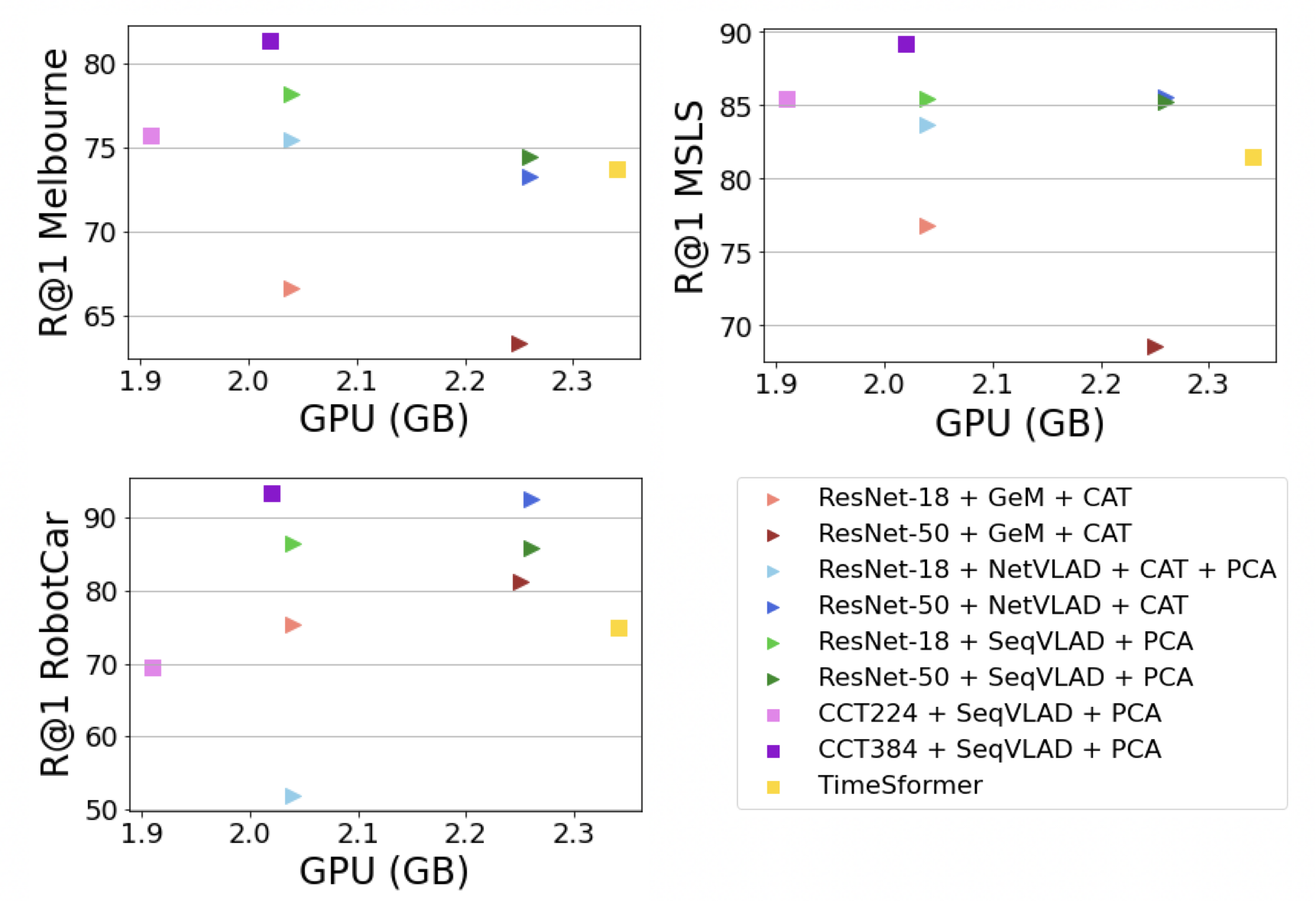}
  \caption{\textbf{Trade-off between GPU memory and recall@1} for best-performing methods}
  \label{fig:gpu_r1}
  \vspace{-0,2cm}
\end{figure}

\myparagraph{Test-time GPU memory requirements} This can be one of the main constraints when choosing a neural network for a real-world application, especially when the descriptors need to be computed on an embedded system.
From \cref{fig:gpu_r1} we find that the best GPU memory-recall trade-off is given by SeqVLAD and PCA with backbones CCT224 and CCT384, with the former being more lightweight and the latter achieving higher recalls.
For example, the CCT384 with SeqVLAD and PCA achieves a recall@1 of 89.2 requiring just 2.02 GB of GPU memory, further confirming the potential of both transformers-based CCT and the novel SeqVLAD layer.

\begin{figure}[t!]
  \centering
  \includegraphics[width=0.7\columnwidth]{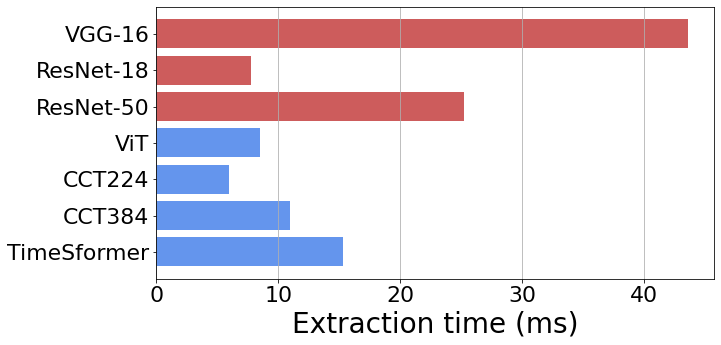}
  \vspace{-0,1cm}
  \caption{\textbf{Extraction time with different backbones} for a sequence of 5 frames. Red and blue bars refer to CNNs and Transformers, respectively. \red{Measured on a NVIDIA Geforce RTX 3090}.
}
  \label{fig:extraction_time}
 \vspace{-0,4cm}
\end{figure}

\myparagraph{Inference time} This factor is crucial when it comes to robotics applications.
In sequence-based VPR, inference time can be divided into two contributions: the extraction time, i.e., how long it takes to extract descriptors for each frame/sequence, and the matching time, which is the delay in retrieving the best match for the query sequence.
Extraction time depends mainly on the backbone of choice, whereas the eventual fusion/aggregation is faster at least one order of magnitude.
For different backbones, extraction time is shown in \cref{fig:extraction_time}, from which we can see that even the best performing backbones require around 10 milliseconds to extract descriptors for a sequence.
However, while extraction time introduces a fixed delay, matching time can quickly represent the bottleneck as the dataset increases in size, given that its complexity linearly depends on the number of images in the database.
In this regard, sequence matching and sequential descriptors rely on two different matching methods: while sequence matching requires a per-frame distance computation, making its complexity $O(N_{db} \cdot \textit{SL} \cdot D)$, where $N_{db}$ is the size of the database, $\textit{SL}$ the sequence length, and $D$ the descriptors dimension.
Methods such as HVPR \cite{Garg-2021-seqNet} propose to use a kNN to find a shortlist of top-K candidates on which to compute the sequential matching leading to a considerable computational speedup, of the order of $O(N_{db} \cdot D + K \cdot \textit{SL} )$.
Instead, sequential descriptors rely solely on a kNN search, of complexity $O(N_{db} \cdot D)$, when considering a number of neighbours $k$ such that $k \ll D$.
As an example, using descriptors of dimension 4096 and a database of 50k, kNN takes around 10 milliseconds. Given their linear relation, a 100 times increase in database size would result in a matching time of 1 second, which would result in extraction time being negligible.
However, sequential descriptors can potentially harness speed increases by taking advantage of the vast literature on approximate kNN techniques that provides considerable speedups at the cost of irrelevant drops in recall, with methods such as inverted file indexes \cite{Sivic-2003},
inverted multi index \cite{BabenkoL12} or hierarchical navigable small world graphs \cite{Malkov2020EfficientAR}.

\myparagraph{Memory footprint} For achieving an efficient retrieval, all database descriptors should be kept in RAM memory, which can be quantified as $N_{db} \times D$.
Considering this, it becomes apparent that the dimensionality of the descriptor (see \cref{tab_all}) can be just as crucial as the recall, given that it linearly increases both memory requirements and matching time.

\begin{table}[t!]
\caption{Robustness to the inversion of the frames.}
\vspace{-0,1cm}
\label{tab_reverseframes}
\begin{adjustbox}{width=.9\columnwidth,center}
\begin{tabular}{cccccc}
\toprule
Method & Backbone & Dim. & Forw. & Back. & Diff \\
\midrule
SeqSLAM* \cite{seqslam}                  & VGG-16    & 4096 & 45.9 & 22.9 & -50 \% \\
Delta D.*\cite{Garg-2020-deltaDescr}     & VGG-16    & 4096 & 43.0 & 11.7 & -73 \% \\
HVPR \cite{Garg-2021-seqNet}             & VGG-16    & 4096 & 51.0 & 28.5 & \underline{-44} \% \\
SeqMatchNet \cite{Garg2021SeqMatchNetCL} & VGG-16    & 4096 & 44.8 & 24.2 & -46 \% \\
\midrule
\midrule
GeM + CAT \cite{Warburg-2020-msls}                                & ResNet-18 & 1280 & 76.8 & 67.8 & -12 \% \\
\textbf{NetVLAD + CAT + PCA}                      & ResNet-18 & 4096 & 83.7 & 79.9 & \underline{-5} \%  \\
\midrule
SeqNet \cite{Garg-2021-seqNet}           & VGG-16    & 4096 & 50.1 & 42.0 & -16 \% \\
\textbf{NetVLAD + FC}                      & ResNet-18 & 4096 & 68.5 & 65.1 & -5 \% \\
\textbf{SeqVLAD + PCA}                            & ResNet-18 & 4096 & 85.5 & 85.2 & -0.4 \% \\
\textbf{SeqVLAD + PCA}                            & CCT384    & 4096 & 89.2 & 89.2 & \underline{\textbf{0.0}} \% \\
\midrule
\textbf{TimeSformer}                              & -         & 768 & 81.5 & 81.5 & \underline{\textbf{0.0}} \%\\
\bottomrule
\end{tabular}
\end{adjustbox}

\vspace{-0.5cm}
\end{table}

\myparagraph{Frame ordering}
This section aims to support the importance of developing sequence-aware methods with experimental evidence. Unlike commonly used datasets like Oxford-Robotcar, where the sequences direction of query and the database follow the same direction, this may not be the case in an actual deployed application. Therefore, it is of paramount importance to have models able to localize a given sequence even if it is encountered with reverse frame ordering with respect to the database.
An obvious workaround would be to include in the database each sequence twice, in both directions; however, this would be a na\"ive solution that would consistently drive up inference time and memory requirements for both sequence-matching and descriptor-based methods.

To shed light on the robustness of different methods to having sequences in different directions, we invert database sequences in the MSLS test set and evaluate previously trained methods on this new dataset.
The results are reported in \cref{tab_reverseframes}, 
which shows that sequence matching is far from ideal for this setting, as it relies on the assumption of fixed direction and speed among the query and database sequences.
Late fusion techniques take a hit in terms of performance, which is easily justified given the nature of the CAT descriptor, which intrinsically incorporates the notion of frame order.
Using the intermediate fusion approach, the FC reduces the recall drop, unlike the simple concatenation, as it can collect features from different frames, but overall it remains a sub-optimal choice. 
Contrarily, SeqVLAD, which already achieves the best performances in the standard setting, proves its value in capturing meaningful representations for localization independently from the frame order.
In the case of early fusion, the TimeSformer shows the same capability in extracting global representations depending only on the multi-view information contained in the frames and not frame-ordering. Like SeqVLAD, this method does not suffer performance hits; however, following the trend highlighted in Tab.~\ref{tab_all}, it achieves lower performance than SeqVLAD.

\begin{table}[t!]
\caption{Flexibility to variations in test-time sequence length (SL). }
\label{tab_seqlen}
  \resizebox{\columnwidth}{!}{
\begin{tabular}{cccccccc}
\toprule
Method & Backbone & Descr. Dim. & \red{SL 1} & \red{SL 3} & SL 5 & SL 10 & SL 15 \\
\midrule
GeM + CAT \cite{Warburg-2020-msls} 			& ResNet-18 & 256 $\cdot$ SL   	& 63.7 & 74.6 & 76.8 & 79.8 & 82.2 \\
GeM + CAT \cite{Warburg-2020-msls} 			& ResNet-50 & 1024 $\cdot$ SL 	& 59.6 & 67.3 & 68.6 & 70.4 &  72.2\\
NetVLAD + CAT \cite{Warburg-2020-msls} 	& ResNet-18 & 16384 $\cdot$ SL 	& 74.4 & 82.2 & 84.3 & 86.2 & 86.9 \\
NetVLAD + CAT \cite{Warburg-2020-msls} 	& ResNet-50 & 65536 $\cdot$ SL 	& \underline{76.0} &\underline{83.1} &  \underline{85.6} & \underline{88.6} & \underline{90.2} \\
\textbf{SeqPool + CAT} 									& CCT224 		& 384 $\cdot$ SL  	& 58.4 & 72.7 & 74.2 & 76.2 & 75.8\\
\textbf{SeqPool + CAT} 									& CCT384 		& 384 $\cdot$ SL  	& 62.2 & 74.6 & 77.8 & 79.9 & 78.8\\
\midrule
\textbf{SeqVLAD + PCA} & ResNet-18 & 4096 & 75.0 & 83.7 & 85.5 & 89.3 & 92.7 \\
\textbf{SeqVLAD + PCA} & ResNet-50 & 4096 & 74.2 & 83.6 & 85.2 & 88.7 & 91.8\\
\textbf{SeqVLAD + PCA} & CCT224    & 4096 & 72.2 & 83.5 & 85.5 & 87.8 & 91.2\\
\textbf{SeqVLAD + PCA} & CCT384    & 4096 & \textbf{\underline{78.2}} & \textbf{\underline{87.6}} & \textbf{\underline{89.2}} & \textbf{\underline{92.1}} & \textbf{\underline{95.2}} \\
\midrule
\textbf{TimeSformer}   &    -      &  768 & \underline{62.2} & \underline{78.8} & \underline{81.5} & \underline{86.3} & \underline{89.9}\\
\bottomrule
\end{tabular}
}
\vspace{-0.5cm}
\end{table}

\myparagraph{Robustness to variations of sequence length at test time}
Several applications of sequence VPR could require the underlying descriptors' extraction to work on sequences of variable length.
For this purpose, the neural network should possess the flexibility of taking as input sequences with a different number of frames without requiring architectural changes.
In this section, we investigate how sequential descriptors that possess this flexibility (namely CAT, SeqVLAD, and TimeSformer), perform when tested with different sequence lengths from those used at training time (i.e., length of 5 frames), and we report the results in \cref{tab_seqlen}.

Methods like TimeSformers and SeqVLAD (with or without PCA) produce descriptors of fixed dimension, which allow for sequences of different lengths to be directly comparable to each other, whereas a simple concatenation of single-frame descriptors intrinsically ties the dimension of the descriptors to the length of the sequence.
From the results in \cref{tab_seqlen}, we can observe a common trend that increasing the length of the sequence results in an increase in recall, confirming previous findings from \cite{Garg2021SeqMatchNetCL}.
This is due to two factors: i) the descriptors are built on more frames, making them more informative and ii) due to the nature of the seq2seq task \cite{Warburg-2020-msls}, longer sequences make it easier for at least one frame to overlap, given that they span over a longer distance.

\section{CONCLUSIONS}
The aim of this work was to provide a new perspective on sequence-based VPR, moving past the na\"ive adaptation of single images methods and proposing novel solutions tailored for sequences.
We have provided a categorization of the alternatives and a thorough benchmark of results on multiple datasets.
Among these new solutions, we have demonstrated the viability of recently proposed Transformer-based architectures, and we have proposed a new layer, SeqVLAD, that is pluggable on top of both kinds of backbones. We show that SeqVLAD sets a new state-of-the-art while reducing computational cost and inference time w.r.t previous methods.
We believe that these novel contributions will pave the way for better sequence-aware methods that fully exploit sequential information. We plan to continue investigating early fusion methods and their robustness under heavy viewpoint changes and domain shifts.




\section*{ACKNOWLEDGMENTS}
We acknowledge the CINECA award under the ISCRA initiative, for the availability of high performance computing resources and support.
Also, computational resources were provided by HPC@POLITO, a project of Academic Computing within the Department of Control and Computer Engineering at the Politecnico di Torino (http://www.hpc.polito.it). This work was supported by CINI.

\bibliographystyle{IEEEtran}
\bibliography{bibliography}

\end{document}